%% file: acl_latex.tex
\pdfoutput=1

\documentclass[11pt]{article}

\usepackage[preprint]{acl}

\usepackage{times}
\usepackage{latexsym}
\usepackage{defs}
\usepackage{booktabs}
\usepackage{dsfont}
\usepackage{enumitem}
\usepackage{float}

\usepackage[T1]{fontenc}

\usepackage[utf8]{inputenc}

\usepackage{microtype}

\usepackage{inconsolata}

\usepackage{graphicx}

\title{Addressing Data Imbalance in Transformer-Based Multi-Label Emotion Detection with Weighted Loss}

\author{
  Xia Cui \\
  Manchester Metropolitan University \\
  \texttt{x.cui@mmu.ac.uk} \\
  }

\begin{document}
\maketitle

\begin{abstract}
This paper explores the application of a simple weighted loss function to Transformer-based models for multi-label emotion detection in SemEval-2025 Shared Task 11~\citep{Muhammad:SemEval:2025}. Our approach addresses data imbalance by dynamically adjusting class weights, thereby enhancing performance on minority emotion classes without the computational burden of traditional resampling methods. We evaluate BERT, RoBERTa, and BART on the BRIGHTER dataset, using evaluation metrics such as Micro F1, Macro F1, ROC-AUC, Accuracy, and Jaccard similarity coefficients. The results demonstrate that the weighted loss function improves performance on high-frequency emotion classes but shows limited impact on minority classes. These findings underscore both the effectiveness and the challenges of applying this approach to imbalanced multi-label emotion detection.

\end{abstract}

\input{input_scripts/introduction}
\input{input_scripts/background}
\input{input_scripts/methods}
\input{input_scripts/experiments}

\input{input_scripts/results}
\input{input_scripts/conclusions}

\section*{Acknowledgments}
We would like to thank the anonymous reviewers for their valuable advice and suggestions.

\bibliography{emotion}

\appendix
\input{input_scripts/appendix}

\end{document}

%% file: input_scripts/introduction.tex
\section{Introduction}
Emotions conveyed through language can manifest via facial expressions, speech, and written text. Unlike facial expressions, which can display a wide spectrum of emotions, or speech, which can express multiple emotions while retaining the same verbal content, emotions in text are particularly challenging to analyse and interpret~\citep{Hancock:SIGCHI:2007}. Textual emotions are especially difficult to analyze due to their subtlety, complexity, and inherent ambiguity in how emotions are expressed. For machines, detecting emotions in text is a particularly formidable task, given the nuanced and often context-dependent nature of emotional language~\citep{pekrun:DP:2022, Mohammad:CL:2022}. Developing an artificial intelligence (AI) system capable of identifying emotional content is even more challenging, as the machine’s role is primarily to support human interpretation, offering insights rather than definitive conclusions—especially given the subjectivity of emotional perception and the likelihood of disagreement among human interpreters~\citep{Schuff:ACL:2017}. This task becomes even more complicated with short texts, which often communicate a blend of emotions simultaneously, making it a critical area of study for advancing both AI and emotional intelligence research.

This paper focuses on developing a system capable of identifying the presence of one or more emotions within short texts as a multi-label classification problem, aiming to advance the accuracy and utility of automated emotion detection. To achieve this, we develop a Transformer-based model, a class of architectures that has recently gained prominence in Natural Language Processing (NLP) for tasks such as sentiment analysis and machine translation. While prior research has extensively explored traditional machine learning (ML) and deep learning (DL) approaches, including Convolutional Neural Networks (CNNs) and Long Short-Term Memory (LSTM) networks~\citep{Zhang:IEEETKDE:2014,Wang:ACL:2016}, this work focuses on the potential of Transformers to address the unique challenges of multi-label emotion detection.

Data imbalance is a common challenge in classification tasks, often leading to poor performance in detecting minority classes. Traditional approaches to address this issue typically involve resampling techniques, such as oversampling minority classes or undersampling majority classes~\citep{Tsoumakas:DKDH:2010,Bach:IS:2017}. However, in multi-label classification tasks, where a single instance can simultaneously belong to both minority and majority classes, these methods become less effective. Resampling in such scenarios complicates the mapping of instances to multiple output labels, as increasing or decreasing the occurrence of one label may inadvertently affect others~\citep{Zhang:IEEETKDE:2014,Charte:Neuro:2015}. In this paper, we investigate the application and impact of class weighting as an alternative strategy for handling data imbalance in multi-label classification, particularly when applied to Transformer-based models.

A key list of our contributions is summarised as follows:
\begin{itemize}[noitemsep,topsep=0pt]
    \item We propose a simple weighted loss function to address data imbalance in multi-label emotion detection using Transformer-based models.
    \item We evaluate three widely used Transformer-based models: BERT, RoBERTa and BART for multi-label emotion detection.
    \item We show that the weighted loss function (+w) enhances performance across all metrics and emotion classes for BERT and BART, while in RoBERTa, it leads to slight underperformance in Micro F1.
\end{itemize}
 
The source code for this paper is publicly available on GitHub\footnote{\url{https://github.com/summer1278/semeval2025-task11}}.

%% file: input_scripts/background.tex
\section{Background}


Significant amounts of studies have been carried out on automatic emotion recognition to help the process of manually checking emotions for various purposes.
Previous studies have explored various machine learning and deep learning approaches to address the challenges of capturing multiple emotions in concise text. \citet{Zhang:IEEETKDE:2014} proposed a framework using binary relevance and classifier chains to handle multi-label classification, demonstrating its effectiveness on social media datasets. \citet{Wang:ACL:2016} proposed leveraging CNN and LSTM to tackle emotion detection, highlighting the potential of deep learning models in capturing complex emotional cues in text.
Transformer-based models such as BERT, RoBERTa, and BART have significantly advanced the field of emotion detection by leveraging their ability to capture contextual relationships in text. BERT (Bidirectional Encoder Representations from Transformers)~\citep{Devlin:BERT:2018} introduced a bidirectional attention mechanism, enabling it to understand the context of words from both left and right, which is particularly useful for detecting nuanced emotional cues in text. RoBERTa (Robustly Optimized BERT Pretraining Approach)~\citep{Liu:roberta:2019} builds on BERT by optimizing its pretraining process, using larger datasets and longer training times, resulting in improved performance on emotion classification tasks. BART (Bidirectional and Auto-Regressive Transformers)~\citep{Lewis:CoRR:2019:BART}, on the other hand, combines bidirectional and autoregressive pretraining objectives, making it effective for both understanding and generating emotionally rich text. These models or variants have been widely adopted in emotion detection due to their ability to handle complex linguistic patterns and their state-of-the-art performance on benchmark datasets~\citep{Ribeiro:ACL:2020,Zhang:EMNLP:2020,Muhammad:BRIGHTER:2025}.

Many methods fail to address label imbalance, leading to biased models~\citep{Zhang:IEEETKDE:2014,Yang:ACL:2019}. \citet{Zhang:EMNLP:2020} proposed a multimodal Transformer-based approach for multi-label emotion detection, modeling both modality and label dependencies. However, emotions are not always correlated and can conflict (e.g., \textit{joy} and \textit{anger}), making dependency assumptions problematic. Their model mitigates imbalance with a weighted loss function and a conditional set generation mechanism, but is computationally expensive due to beam search and permutation-based training. 
In this paper, we employ a similar weighted loss function that adjusts class weights directly, reducing complexity and resource demands, to the Transformer-based models.

%% file: input_scripts/methods.tex
\section{Methods}

We consider the emotion detection task as a multi-label classification problem.
Multi-label classification involves assigning multiple labels to each input instance. Section~\ref{sec:problem-def}, we provide the definition of the problem with notation. Unlike traditional classification tasks, where each input is associated with a single label, multi-label classification requires the model to predict a subset of labels from the label space.
Section~\ref{sec:transformer-backbone}, we introduce the Transformer-based architecture including critical components and how we apply instance weighting in the loss function for the defined multi-label classification problem.

\subsection{Problem Definition}\label{sec:problem-def}
Let the dataset consist of $N$ samples, where each sample is represented as a text input $x_i$ (e.g., a sentence or paragraph). These inputs are preprocessed and transformed into numerical representations, forming a feature matrix $X \in \mathbb{R}^{N \times d}$, where $d$ is the feature dimensionality. The corresponding emotion labels are represented by a multilabel target matrix $Y \in \{0, 1\}^{N \times C}$, where $C$ is the number of possible emotion classes (e.g., "joy", "sadness", "anger", etc.). Each element $Y_{ij}$ indicates whether the $j$-th emotion is expressed in the $i$-th input: 

{\small
\begin{equation}
    Y_{ij} = \begin{cases} 
        1, & \text{if sample } i \text{ contains a emotion label } j, \\
        0, & \text{otherwise}.
    \end{cases}
\end{equation}
}

The goal is to learn a function $f: \mathbb{R}^d \to [0, 1]^C$ that maps an input feature vector $x \in \mathbb{R}^d$ to a probability vector $\hat{y} \in [0, 1]^C$, where each $\hat{y}_j$ represents the predicted probability of label $j$ being relevant.

\subsection{Transformer-based Backbone}\label{sec:transformer-backbone}
We propose a model based on the Transformer architecture~\citep{Vaswani:NIPS:2017}, which employs self-attention mechanisms and positional encoding to effectively capture contextual relationships in text. While the Transformer architecture itself remains unchanged, we introduce a weighted loss function to address the challenge of data imbalance in multi-label classification. Specifically, the weighted loss function dynamically adjusts the contribution of each instance during training based on its label distribution, enabling the model to better handle minority classes without disrupting the relationships between labels. This approach allows us to leverage the strengths of Transformers while mitigating the biases introduced by imbalanced data.

\subsubsection{Cross-Entropy Loss}\label{sec:ce}
Training Transformer-based models for classification tasks typically employs the cross-entropy loss function. This loss quantifies the discrepancy between the predicted probability distribution and the true distribution, serving as a standard objective to optimize the performance of the model.

Let $y$ denote the true target sequence and $\hat{y}$ the predicted sequence. The cross-entropy loss is defined as,
\begin{equation}
    \mathcal{L}_{\mathrm{CE}} = -\frac{1}{N} \sum_{i=1}^N \sum_{j=1}^{C} y_{ij} \log(\hat{y}_{ij}),
\end{equation}
where $N$ is the number of samples in the batch, $C$ is the number of emotion classes, $y_{ij}$ is a binary indicator (1 if the true class for sample $i$ is $j$, 0 otherwise) and $\hat{y}_{ij}$ is the predicted probability for class $j$ for sample $i$.

Binary Cross-Entropy (BCE) loss is employed for multi-label classification tasks. Under the assumption that emotions are independent, the BCE loss is calculated separately for each label, leading to the following formulation of the function:

{\small
\begin{align}
    \mathcal{L} = -\frac{1}{N} \sum_{i=1}^N \sum_{j=1}^C \left[ y_{ij} \log(\hat{y}_{ij}) + (1 - y_{ij}) \log(1 - \hat{y}_{ij}) \right]
    \label{eq:bce}
\end{align}
}
This loss encourages the model to assign high probabilities to the correct labels and low probabilities to the incorrect labels for each instance.

\subsubsection{Class Weights and Label Smoothing}
To enhance generalization, a common strategy is to incorporate class weights into the loss function~\citep{Ridnik:ICCV:2021}:

{\small
\begin{align}
    \mathcal{L}^\prime = -\frac{1}{N} \sum_{i=1}^{N} \sum_{j=1}^{C} w_j \left[ y_{ij} \log(\hat{y}_{ij}) + (1 - y_{ij}) \log(1 - \hat{y}_{ij}) \right]
\end{align}
}
Here, $w_j$ denotes the weight assigned to the $j$-th class, computed as $w_j = f_j / N$, where $f_j$ is the frequency of the $j$-th class in the dataset, and $N$ is the total number of training instances.
$w_j$ is normalised as $w_j = \max(W) / w_j$, where $W$ represents an array of distributions from $C$ labels in the training dataset.

Additionally, label smoothing is frequently employed, where the traditional one-hot target distribution is replaced with a smoothed version to reduce the overfitting of majority classes:
\begin{equation}
    y_{ij}^{\text{smooth}} = (1 - \epsilon) y_{ij} + \frac{\epsilon}{C}
\end{equation}
where $\epsilon$ is the smoothing parameter. This prevents the model from becoming overconfident in its predictions.

\subsubsection{Prediction}
The predicted probabilities for all samples can be represented as a matrix:
\begin{equation}
    \hat{Y} = \sigma(Z), \quad Z \in \mathbb{R}^{N \times C}
\end{equation}
where $Z$ is the output logits of the model and $\sigma(\cdot)$ is the element-wise sigmoid function, defined as $\sigma(z) = 1/ (1 + e^{-z})$.
In a multi-label classification task, where the model operates as a collection of independent binary classifiers, a probability threshold of $\tau = 0.5$ is typically applied to determine label assignments. However, if all predicted probabilities fall below $\tau$, the label $y_j$ corresponding to the highest logit value $z_j$ (i.e., $y_j = \arg\max_{j \in \{1, \ldots, C\}} z_j$ ) is assigned as the final prediction to mitigate unclassified instances.  

\subsubsection{Training Objective}\label{sec:training-obj}

During training, the model minimizes the BCE loss Eq.~\ref{eq:bce}, while the Macro F1 score (Eq.~\ref{eq:marof1}) is monitored using a development set to select the best model,
\begin{align}
    \theta^* = \arg\min_{\theta} \mathcal{L}(f_\theta(X), Y),
\end{align}
where $\theta$ are the parameters of the model, $f_\theta(X)$ is the output of the model for input $X$ and $f_\theta(X)=\hat{Y}$.

%% file: input_scripts/experiments.tex
\section{Experimental Setup}\label{sec:exp-setup}
The task is framed as a multi-label classification problem. The proposed system is developed using the BRIGHTER dataset~\citep{Muhammad:BRIGHTER:2025}, which includes multi-label emotion annotations across 28 languages. To address the challenge of data imbalance, we focus our experiments on the English subset as a representative example. The proposed approach is designed to be generalizable and can be extended to other languages, such as German, with minimal adaptation.

\subsection{Experimental Data and Preprocessing}
\begin{figure}
    \centering
    \includegraphics[width=0.7\linewidth]{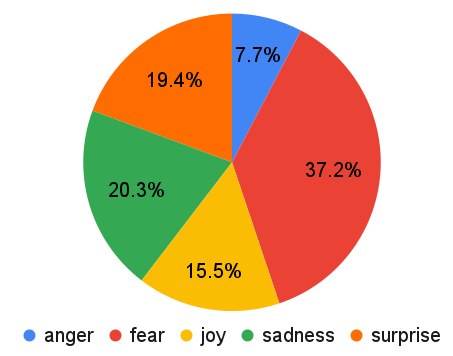}
    \caption{Label distributions of training set in BRIGHTER English Track A dataset.}
    \label{fig:label-dis}
\end{figure}

Each instance is annotated with binary presence labels (1 for positive, 0 for negative) across five emotion classes: anger, fear, joy, sadness, and surprise. Table~\ref{fig:label-dis} illustrates the label distribution for the training set in the BRIGHTER English Track A dataset. Since the test set labels were not accessible during the evaluation phase, we used the official development split (116) as a test set. The original training set was divided into 70\% (1937) for training and 30\% (831) for development. No additional training datasets were introduced to boost the perfomrance.

For data preprocessing, we employed the \texttt{AutoTokenizer}\footnote{\url{https://huggingface.co/transformers/v4.7.0/model_doc/auto.html\#transformers.AutoTokenizer}}, which tokenizes the input text and generates corresponding \texttt{attention\_mask} and \texttt{input\_ids} for the model.
For consistency and reproducibility, we utilize \texttt{AutoTokenizer} with default settings for all models. The tokenization process follows model-specific encoding methods, including WordPiece for BERT-based models and Byte-Pair Encoding for RoBERTa and BART. Sequences are automatically padded and truncated to a fixed length, ensuring uniform input across all models. For uncased models, all text is lower-cased, whereas cased models preserve the original casing. Additionally, special tokens are inserted according to each model’s pretrained configuration. We do not apply any additional text normalization, stemming, or lemmatization beyond what is internally handled by the tokenizer.

\subsection{Hyperparameter Tuning}
To identify optimal baseline hyperparameters across all models, we employ \textsc{Optuna}~\citep{Akiba:SIGKDD:2019:optuna}, a Bayesian optimization framework, to maximize the Macro F1 score (Section~\ref{sec:training-obj}), which is a class-imbalance-resistant metric particularly suitable for multi-label emotion detection. 
Our experiments use the \texttt{BERT-base-uncased} model checkpoint\footnote{\url{https://huggingface.co/google-bert/bert-base-uncased}} as the foundation. The hyperparameter search executes 10 optimization trials, with each trial training on the training set and evaluating on the development set to ensure robust generalization. 
We use \textsc{Ray Tune}~\citep{Liaw:2018:raytune} to distribute these trials across available computing resources.
The configuration yielding the best performance is selected for the final training phase. The optimized hyperparameters obtained through tuning were: learning rate ($\eta$) = $2.45 \times 10^{-5}$, batch size = 8 and number of epochs = 3. Additional results are provided in Appendix Section~\ref{sec:appendix-trails}.

\subsection{Implemention Details}
All experiments are conducted using Google Colab’s T4 GPU server (NVIDIA T4 GPU with 16GB of VRAM)\footnote{\url{https://colab.research.google.com/}}, leveraging the HuggingFace \texttt{transformers} library with PyTorch as the backend. The models are trained using AdamW optimiser~\citep{Loshchilov:AdamW:2017} with a linear learning rate scheduler, and a batch size of 8 is used throughout the experiments. To prevent overfitting, early stopping is applied based on validation performance. Evaluation is performed using standard classification metrics implemented via the \texttt{sklearn.metrics} module\footnote{\url{https://scikit-learn.org/stable/api/sklearn.metrics.html}}. 

\subsection{Evaluation Metrics}
We evaluate the performance of this multi-label classification task using four metrics:
F1 scores, ROC-AUC score, Accuracy (Acc) and Jaccard similarity coefficients.

The F1 score $F_1$ for a binary classification task is computed using the number of True Positives (TP), False Positives (FP) and False Negatives (FN):
\begin{align}
    F_1 = \frac{\mathrm{TP}}{\mathrm{TP}+\frac{1}{2}\cdot(\mathrm{FP}+\mathrm{FN})}\label{eq:f1}
\end{align}

Considering a multiclass problem under the one-vs-rest strategy, having $C$ classes, Micro F1 is computed using the total number of TP, FP and FN across all classes:

{\small
\begin{align}
    \mathrm{Micro}F_1 = \frac{ \sum_{j=1}^{C} {\mathrm{TP}_j}}{\sum_{j=1}^{C} {\mathrm{TP}_j}+\frac{1}{2}\cdot\(\sum_{j=1}^{C} {\mathrm{FP}_j+\sum_{j=1}^{C} \mathrm{FN}_j}\)}
\end{align}
}
Whereas Macro F1 takes an average of class-wise $F_1$ across all classes:
\begin{align}
    \mathrm{Macro} F_1 = \frac{\sum^C_{j=1} F_1(j)}{C}\label{eq:marof1}
\end{align}
The Weighted F1 score is a metric commonly used in multi-label classification tasks, particularly in scenarios involving imbalanced datasets. It calculates the F1 score for each class (which can be considered a binary classification problem to be computed by Eq.~\ref{eq:f1}) and takes the average, weighted by the number of true instances (support) for each class. This ensures that classes with more instances have a larger impact on the final score, which is especially important when dealing with imbalanced classes. It is given by:
\begin{align}
   \mathrm{Weighted}F1 = \frac{\sum_{j=1}^{C} w_j \cdot F_1(j)}{\sum_{j=1}^{C} w_j} 
\end{align}
where  $C$  is the total number of classes,  $F_1(j)$ is the F1 score for class \textit{j}, $w_i$ is the support (the number of true instances) for class $j$.
This formulation ensures that classes with more instances (higher support) have a proportionally larger effect on the final weighted F1 score. In our experiments, we use this metric to evaluate the overall performance of the models while accounting for class imbalance.
Another alternative method to calculate the F1 score for a multiclass problem is to average the score across $N$ instances. This approach is known as the sample-averaged F1 score:
\begin{align}
    \mathrm{Sample} F_1 = \frac{\sum^N_{i=1} F_1(i)}{N}
\end{align}

The Receiver Operating Characteristic (ROC) curve plots the True Positive Rate (TPR) against the False Positive Rate (FPR) across various thresholds, where TPR = TP / (TP + FN) and FPR = FP / (FP + TN). The Area Under the Curve (AUC) represents the likelihood that a randomly selected positive instance is ranked higher than a randomly selected negative one. A higher AUC indicates better model performance.

The Jaccard similarity coefficient measures the similarity between predicted labels $\mathrm{\mat{y}_{pred}}$ and true labels $\mathrm{\mat{y}_{true}}$, calculated as the size of their intersection divided by the size of their union:
\begin{align}
    \mathrm{Jaccard} = \frac{|\mathrm{\mat{y}_{pred}}\cap \mathrm{\mat{y}_{true}}|}{|\mathrm{\mat{y}_{pred}}\cup \mathrm{\mat{y}_{true}}|}
\end{align}

%% file: input_scripts/results.tex
\section{Results}\label{sec:results}

\begin{table}[]
\caption{Performance on Transformer-based models with or without weighted loss function (+w). The best results are bolded.}
\label{tab:transformer-results}
\resizebox{0.496\textwidth}{!}{%
\begin{tabular}{@{}lccccc@{}}
\toprule
&\textbf{$\mathrm{Micro}F_1$} & \textbf{$\mathrm{Macro}F_1$}& ROC-AUC & Acc & Jaccard \\ \midrule
BERT       & 0.7095                                 & 0.6859                                 & 0.7927                                & 0.3621                                & 0.5776                               \\
BERT+w & \textbf{0.7198}                                 & \textbf{0.7008 }                                & \textbf{0.8016}                                & \textbf{0.3966 }                               & \textbf{0.5991 }                              \\ \midrule
RoBERTa     & \textbf{0.7282}                                 & 0.7162                                 & 0.8116                                & 0.3707                                & 0.5934                               \\
RoBERTa+w & 0.7268                                 & \textbf{0.7184 }                                &\textbf{ 0.8127 }                               & \textbf{0.3793}                                & \textbf{0.6013 }                              \\ \midrule
BART        & 0.6961                                 & 0.6803                                 & 0.7837                                & 0.3707                                & 0.5668                               \\
BART+w & \textbf{0.7321}                                 & \textbf{0.7136 }                                & \textbf{0.8141}                                & \textbf{0.4138 }                               & \textbf{0.6114 }                              \\ \bottomrule
\end{tabular}
}
\end{table}
Table~\ref{tab:transformer-results} presents the results of five major evaluation metrics for three Transformer-based models: BERT, RoBERTa and BART. Due to computational constraints, we use the base versions of these models, leveraging pre-trained checkpoints available on HuggingFace Hub\footnote{\url{https://huggingface.co/models}}.
To ensure a fair comparison, all models are fine-tuned under identical conditions, including consistent hyperparameters, epochs, and optimization settings, isolating the impact of model architectures (Section~\ref{sec:exp-setup}).

The comparative evaluation of BERT, BART, and RoBERTa, both with and without the weighted loss function (+w), reveals distinct patterns in their handling of multi-label emotion detection. BERT+w and BART+w show notable improvements, particularly in Weighted F1 (BERT: from 0.7115 to 0.7228 and BART: from 0.6993 to 0.7350), demonstrating enhanced detection of both majority and minority emotions. BART+w exhibits the most significant gains, with Micro F1 increasing from 0.6961 to 0.7321, reflecting stronger overall classification robustness. In contrast, RoBERTa+w shows only marginal changes, with a slight decrease in Weighted F1 (from 0.7298 to 0.7270) and near-identical Micro F1 (from 0.7282 to 0.7268), suggesting that RoBERTa is inherently well-optimized for handling class imbalance. Further analysis (Table~\ref{tab:vs-label} and Table~\ref{tab:vs-f1}) reveals that improvements in Macro F1 and Sample F1 are primarily driven by gains in high-frequency classes, with limited or no improvements in most emotion classes (4 out of 5). This indicates that the weighted loss function disproportionately benefits high-frequency classes, potentially masking stagnation in minority emotion detection. Additionally, RoBERTa's performance suggests that it is less sensitive to data imbalance than BERT and BART, likely due to its more robust pretraining, enabling it to outperform these models in the base configuration. Further analysis on RoBERTa's higher precision and recall for minority classes without reweighting is provided in Appendix Section~\ref{sec:appendix-more-results}. In contrast, BART is highly sensitive to data imbalance, resulting in the lowest F1 scores among the three models. However, with the +w variation, BART’s performance improves significantly, surpassing the other models in 4 out of 5 major evaluation metrics, with only a slight drop in Macro F1 or closely matching their performance.

Additionally, we observe that \textit{joy} and \textit{surprise} tend to have overlapping misclassifications, possibly due to contextual ambiguity. This issue highlights a potential limitation in current text-based emotion representations, where fine-grained distinctions between similar emotions remain challenging for Transformer-based models. 

\begin{table}[]
\centering
\caption{Class-wise Macro F1 scores for all models.}
\label{tab:vs-label}
\resizebox{0.43\textwidth}{!}{%
\begin{tabular}{@{}lccccc@{}}
\toprule
Class &
  anger &
  fear &
  joy &
  sadness &
  surprise \\ \midrule
BERT & 0.5806 & 0.7482 & \textbf{0.6667} & 0.7385 & 0.6957 \\
BERT+w & \textbf{0.6471} & \textbf{0.7571} & 0.6296 & \textbf{0.7429} & \textbf{0.7273} \\\midrule
RoBERTa & \textbf{0.7179} & \textbf{0.7714} & \textbf{0.7368} & \textbf{0.7317} & 0.6230 \\
RoBERTa+w & 0.7000 & 0.7534 & \textbf{0.7368} & 0.7250 & \textbf{0.6769} \\ \midrule
BART &
  0.6452 &
  0.7413 &
  0.6792 &
  0.6087 &
  0.7273 \\
BART+w &
  \textbf{0.6875} &
  \textbf{0.7943} &
  \textbf{0.7037} &
  \textbf{0.6389} &
  \textbf{0.7436} \\ \bottomrule
\end{tabular}%
}
\end{table}

\begin{table}[t]
\centering
\caption{F1 score variations (macro, weighted, micro, and samples averaging) for all models.}
\label{tab:vs-f1}
\resizebox{0.46\textwidth}{!}{%
\begin{tabular}{@{}lcccc@{}}
\toprule
       &\textbf{$\mathrm{Micro}F_1$} & \textbf{$\mathrm{Macro}F_1$} & \textbf{$\mathrm{Weighted}F_1$} & \textbf{$\mathrm{Sample}F_1$} \\ \midrule
BERT & 0.7095 & 0.6859 & 0.7115 & 0.6514 \\
BERT+w & \textbf{0.7198} & \textbf{0.7008} & \textbf{0.7228} & \textbf{0.6625} \\ \midrule
RoBERTa &\textbf{ 0.7282} & 0.7162 & \textbf{0.7298 }& 0.6644 \\
RoBERTa+w & 0.7268 & \textbf{0.7184} & 0.7270 & \textbf{0.6705}\\ \midrule
BART   & 0.6961                      & 0.6803                      & 0.6993                         & 0.6333                       \\
BART+w & \textbf{0.7321}             & \textbf{0.7136}             & \textbf{0.7350}                & \textbf{0.6766}              \\ \bottomrule
\end{tabular}%
}
\end{table}

%% file: input_scripts/conclusions.tex
\section{Conclusion}

This paper contributes to the field of multi-label emotion detection by proposing a simplified weighted loss function that mitigates the effects of data imbalance in Transformer-based models. We demonstrate the application of the weighted loss function (+w) improves performance for BERT and BART models, challenges remain in detecting minority emotions. The findings suggest that future work could expand on this approach, particularly by exploring the impact of the proposed method across different languages and datasets.


%% file: input_scripts/appendix.tex
\clearpage

\section{Hyperparameter Tuning Trails}\label{sec:appendix-trails}
Each trail in \textsc{Ray Tune} is an independent run with a unique set of hyperparameters.
The \textsc{Optuna} optimization over 10 trials revealed several key insights: (a) Moderate learning rates ($2.45 \times 10^{-5}$ to $5.79 \times 10^{-5}$) achieved optimal performance, with the best configuration (Trial 4) yielding a Macro F1 score of 0.724; (b) Smaller batch sizes (4-16) outperformed larger ones, suggesting the importance of more frequent gradient updates; (c) Training stability was maintained with 3-5 epochs, avoiding overfitting while achieving convergence. Two trials were pruned early due to poor validation performance. The optimal configuration balanced learning dynamics (batch size 8 and 3 epochs) with precise weight updates ($\eta=2.45 \times 10^{-5}$), demonstrating the effectiveness of Bayesian optimization for transformer fine-tuning.

\begin{table}[h]
\centering
\caption{Optuna hyperparameter optimization results (top 4 trails). $\eta$ denotes the learning rate.}
\label{tab:optuna_results}
\resizebox{0.48\textwidth}{!}{%
\begin{tabular}{lcccc}
\toprule
\#trial & $\eta$ & batch size & \#epochs & Macro$F_1$ \\
\midrule
4 (best) & $2.45 \times 10^{-5}$ & 8 & 3 & 0.7239 \\
2 & $4.14 \times 10^{-5}$ & 4 & 3 & 0.7105 \\
3 & $5.79 \times 10^{-5}$ & 4 & 4 & 0.7162 \\
5 & $3.39 \times 10^{-5}$ & 16 & 5 & 0.6911 \\
\bottomrule
\end{tabular}%
}
\end{table}


\section{Discussion on Classification Reports}\label{sec:appendix-more-results}
Due to page limitations, we present the complete classification reports for all models in Table~\ref{tab:classification-report}, including the base versions of BERT, RoBERTa, and BART, along with their weighted loss function (+w) variants. Across all models, precision generally improves with the +w variation, particularly for minority emotions such as \textit{anger} and \textit{surprise}, while recall tends to fluctuate, sometimes decreasing slightly. These results further support the findings discussed in Section~\ref{sec:results}, indicating that the weighted loss function effectively mitigates class imbalance for BERT and BART but offers limited benefits for RoBERTa—likely due to its already strong baseline performance in multi-label classification tasks.

RoBERTa demonstrates notable resilience to data imbalance even without reweighting strategies. We attribute this robustness to its enhanced pretraining methodology, which includes training on a significantly larger corpus, longer training duration, dynamic masking, and the removal of the Next Sentence Prediction (NSP) objective~\citep{Liu:roberta:2019}. These improvements result in richer and more generalizable contextual representations, which likely contribute to RoBERTa's stable performance across both frequent and minority emotion classes. As shown in our baseline results, RoBERTa achieves relatively high precision and recall for low-frequency classes such as \textit{anger} and \textit{surprise}, even without any rebalancing. In contrast, BERT and BART exhibit more pronounced performance drops for these underrepresented classes. These findings suggest that RoBERTa’s pretraining design inherently mitigates some of the negative effects of label imbalance, making it a particularly strong baseline in imbalanced multi-label emotion detection settings.

\begin{table*}[b!]
\centering
\caption{Classification reports for BERT/BERT+w, BART/BART+w and RoBERTa/RoBERTa+w.}
\label{tab:classification-report}
\resizebox{0.9\textwidth}{!}{%
\begin{tabular}{@{}c|cccc|cccc@{}}
\toprule
\multicolumn{1}{l|}{}    & \multicolumn{4}{c}{BERT}       & \multicolumn{4}{|c}{BERT+w}     \\ \midrule
\multicolumn{1}{r|}{\textbf{}} &
  \textbf{precision} &
  \textbf{recall} &
  \textbf{f1-score} &
  \textbf{support} &
  \textbf{precision} &
  \textbf{recall} &
  \textbf{f1-score} &
  \textbf{support} \\ \midrule
\textbf{anger}          & 0.5625 & 0.6000 & 0.5806 & 15  & 0.6875 & 0.6111 & 0.6471 & 18  \\ 
\textbf{fear}           & 0.8254 & 0.6842 & 0.7482 & 76  & 0.8413 & 0.6883 & 0.7571 & 77  \\ 
\textbf{joy}            & 0.5806 & 0.7826 & 0.6667 & 23  & 0.5484 & 0.7391 & 0.6296 & 23  \\ 
\textbf{sadness}        & 0.6857 & 0.8000 & 0.7385 & 30  & 0.7429 & 0.7429 & 0.7429 & 35  \\ 
\textbf{surprise}       & 0.7742 & 0.6316 & 0.6957 & 38  & 0.7742 & 0.6857 & 0.7273 & 35  \\ \midrule
\textbf{micro avg}      & 0.7216 & 0.6978 & 0.7095 & 182 & 0.7443 & 0.6968 & 0.7198 & 188 \\ 
\textbf{macro avg}      & 0.6857 & 0.6997 & 0.6859 & 182 & 0.7188 & 0.6934 & 0.7008 & 188 \\ 
\textbf{weighted   avg} & 0.7391 & 0.6978 & 0.7115 & 182 & 0.7599 & 0.6968 & 0.7228 & 188 \\ 
\textbf{samples   avg}  & 0.6681 & 0.6997 & 0.6514 & 182 & 0.6782 & 0.7026 & 0.6625 & 188 \\ \bottomrule

\toprule
\multicolumn{1}{l|}{}   & \multicolumn{4}{c|}{RoBERTa}   & \multicolumn{4}{c}{RoBERTa+w}  \\ \midrule
\multicolumn{1}{r|}{\textbf{}} &
  \multicolumn{1}{c}{\textbf{precision}} &
  \multicolumn{1}{c}{\textbf{recall}} &
  \multicolumn{1}{c}{\textbf{f1-score}} &
  \multicolumn{1}{c|}{\textbf{support}} &
  \multicolumn{1}{c}{\textbf{precision}} &
  \multicolumn{1}{c}{\textbf{recall}} &
  \multicolumn{1}{c}{\textbf{f1-score}} &
  \multicolumn{1}{c}{\textbf{support}} \\ \midrule
\textbf{anger}          & 0.8750 & 0.6087 & 0.7179 & 23  & 0.8750 & 0.5833 & 0.7000 & 24  \\
\textbf{fear}           & 0.8571 & 0.7013 & 0.7714 & 77  & 0.8730 & 0.6627 & 0.7534 & 83  \\
\textbf{joy}            & 0.6774 & 0.8077 & 0.7368 & 26  & 0.6774 & 0.8077 & 0.7368 & 26  \\
\textbf{sadness}        & 0.8571 & 0.6383 & 0.7317 & 47  & 0.8286 & 0.6444 & 0.7250 & 45  \\
\textbf{surprise}       & 0.6129 & 0.6333 & 0.6230 & 30  & 0.7097 & 0.6471 & 0.6769 & 34  \\ \midrule
\textbf{micro avg}      & 0.7841 & 0.6798 & 0.7282 & 203 & 0.8011 & 0.6651 & 0.7268 & 212 \\
\textbf{macro avg}      & 0.7759 & 0.6779 & 0.7162 & 203 & 0.7927 & 0.6690 & 0.7184 & 212 \\
\textbf{weighted   avg} & 0.8001 & 0.6798 & 0.7298 & 203 & 0.8136 & 0.6651 & 0.7270 & 212 \\
\textbf{samples   avg}  & 0.7105 & 0.6767 & 0.6644 & 203 & 0.7241 & 0.6710 & 0.6705 & 212 \\ \bottomrule

\toprule
\multicolumn{1}{l|}{}   & \multicolumn{4}{c|}{BART}      & \multicolumn{4}{c}{BART+w}     \\ \midrule
\multicolumn{1}{r|}{\textbf{}} &
  \multicolumn{1}{c}{\textbf{precision}} &
  \multicolumn{1}{c}{\textbf{recall}} &
  \multicolumn{1}{c}{\textbf{f1-score}} &
  \multicolumn{1}{c|}{\textbf{support}} &
  \multicolumn{1}{c}{\textbf{precision}} &
  \multicolumn{1}{c}{\textbf{recall}} &
  \multicolumn{1}{c}{\textbf{f1-score}} &
  \multicolumn{1}{c}{\textbf{support}} \\ \midrule
\textbf{anger}          & 0.6250 & 0.6667 & 0.6452 & 23  & 0.6875 & 0.6875 & 0.6875 & 24  \\
\textbf{fear}           & 0.8413 & 0.6625 & 0.7413 & 77  & 0.8889 & 0.7179 & 0.7943 & 83  \\
\textbf{joy}            & 0.5806 & 0.8182 & 0.6792 & 26  & 0.6129 & 0.8261 & 0.7037 & 26  \\
\textbf{sadness}        & 0.6000 & 0.6176 & 0.6087 & 47  & 0.6571 & 0.6216 & 0.6389 & 45  \\
\textbf{surprise}       & 0.7742 & 0.6857 & 0.7273 & 30  & 0.9355 & 0.6170 & 0.7436 & 34  \\ \midrule
\textbf{micro avg}      & 0.7159 & 0.6774 & 0.6961 & 203 & 0.7841 & 0.6866 & 0.7321 & 212 \\
\textbf{macro avg}      & 0.6842 & 0.6901 & 0.6803 & 203 & 0.7564 & 0.6940 & 0.7136 & 212 \\
\textbf{weighted   avg} & 0.7363 & 0.6774 & 0.6993 & 203 & 0.8095 & 0.6866 & 0.7350 & 212 \\
\textbf{samples   avg}  & 0.6523 & 0.6688 & 0.6333 & 203 & 0.7170 & 0.6868 & 0.6766 & 212 \\ \bottomrule
\end{tabular}%
}
\end{table*}

%% file: emotion.bib
@inproceedings{Muhammad:SemEval:2025,
  title = "{S}em{E}val Task 11: Bridging the Gap in Text-Based Emotion Detection",
  author = "Muhammad, Shamsuddeen Hassan and Ousidhoum, Nedjma and Abdulmumin, Idris and Yimam, Seid Muhie and Wahle, Jan Philip and Ruas, Terry and Beloucif, Meriem and De Kock, Christine and Belay, Tadesse Destaw and Ahmad, Ibrahim Said and Surange, Nirmal and Teodorescu, Daniela and Adelani, David Ifeoluwa and Aji, Alham Fikri and Ali, Felermino and Araujo, Vladimir and Ayele, Abinew Ali and Ignat, Oana and Panchenko, Alexander and Zhou, Yi and Mohammad, Saif M.",
  booktitle = "Proceedings of the 19th International Workshop on Semantic Evaluation (SemEval-2025)",
  month = "july",
  year = "2025",
  address = "Vienna, Austria",
  publisher = "Association for Computational Linguistics"
}

@article{Liaw:2018:raytune,
  title={Tune: A Research Platform for Distributed Model Selection and Training},
  author={Liaw, Richard and Liang, Eric and Nishihara, Robert and Moritz, Philipp and Gonzalez, Joseph E. and Stoica, Ion},
  journal={arXiv preprint arXiv:1807.05118},
  year={2018}
}

@inproceedings{Akiba:SIGKDD:2019:optuna,
  title={Optuna: A Next-generation Hyperparameter Optimization Framework},
  author={Akiba, Takuya and Sano, Shotaro and Yanase, Toshihiko and Ohta, Takeru and Koyama, Masanori},
  booktitle={Proceedings of the 25th {ACM} {SIGKDD} International Conference on Knowledge Discovery and Data Mining},
  year={2019}
}

@inproceedings{Hancock:SIGCHI:2007,
  author = {Hancock, Jeffrey T. and Landrigan, Christopher and Silver, Courtney},
  title = {Expressing emotion in text-based communication},
  year = {2007},
  publisher = {Association for Computing Machinery},
  address = {New York, NY, USA},
  booktitle = {Proceedings of the SIGCHI Conference on Human Factors in Computing Systems},
  pages = {929--932}
}

@article{pekrun:DP:2022,
  title={Emotions in reading and learning from texts: Progress and open problems},
  author={Pekrun, Reinhard},
  journal={Discourse Processes},
  volume={59},
  number={1-2},
  pages={116--125},
  year={2022},
  publisher={Taylor \& Francis}
}

@article{Vaswani:NIPS:2017,
  title={Attention is All You Need},
  author={Vaswani, Ashish and Shazeer, Noam and Parmar, Niki and Uszkoreit, Jakob and Jones, Llion and Gomez, Aidan N. and Kaiser, {\L}ukasz and Polosukhin, Illia},
  journal={Advances in Neural Information Processing Systems},
  year={2017}
}

@article{Mohammad:CL:2022,
  title={Ethics sheet for automatic emotion recognition and sentiment analysis},
  author={Mohammad, Saif M.},
  journal={Computational Linguistics},
  volume={48},
  number={2},
  pages={239--278},
  year={2022},
  publisher={MIT Press}
}

@inproceedings{Schuff:ACL:2017,
  title = "Annotation, Modelling and Analysis of Fine-Grained Emotions on a Stance and Sentiment Detection Corpus",
  author = "Schuff, Hendrik and Barnes, Jeremy and Mohme, Julian and Pad{\'o}, Sebastian and Klinger, Roman",
  booktitle = "Proceedings of the 8th Workshop on Computational Approaches to Subjectivity, Sentiment and Social Media Analysis",
  year = "2017",
  address = "Copenhagen, Denmark",
  publisher = "Association for Computational Linguistics",
  pages = "13--23"
}

@inproceedings{Devlin:BERT:2018,
  title = "{BERT}: Pre-training of Deep Bidirectional Transformers for Language Understanding",
  author = "Devlin, Jacob and Chang, Ming-Wei and Lee, Kenton and Toutanova, Kristina",
  booktitle = "Proceedings of the 2019 Conference of the North American Chapter of the Association for Computational Linguistics",
  year = "2019",
  pages = "4171--4186"
}

@article{Liu:roberta:2019,
  title={RoBERTa: A robustly optimized BERT pretraining approach},
  author={Liu, Yinhan and Ott, Myle and Goyal, Naman and Du, Jingfei and Joshi, Mandar and Chen, Danqi and Levy, Omer and Lewis, Mike and Zettlemoyer, Luke and Stoyanov, Veselin},
  journal={arXiv preprint arXiv:1907.11692},
  year={2019}
}

@article{Lewis:CoRR:2019:BART,
  author={Lewis, Mike and Liu, Yinhan and Goyal, Naman and Ghazvininejad, Marjan and Mohamed, Abdelrahman and Levy, Omer and Stoyanov, Veselin and Zettlemoyer, Luke},
  title={BART: Denoising Sequence-to-Sequence Pre-training for Natural Language Generation, Translation, and Comprehension},
  journal={CoRR},
  volume={abs/1910.13461},
  year={2019}
}

@ARTICLE{Zhang:IEEETKDE:2014,
  author={Zhang, Min-Ling and Zhou, Zhi-Hua},
  title={A Review on Multi-Label Learning Algorithms},
  journal={IEEE Transactions on Knowledge and Data Engineering},
  year={2014},
  volume={26},
  number={8},
  pages={1819--1837}
}

@inproceedings{Wang:ACL:2016,
  title = "Dimensional Sentiment Analysis Using a Regional {CNN}-{LSTM} Model",
  author = "Wang, Jin and Yu, Liang-Chih and Lai, K. Robert and Zhang, Xuejie",
  booktitle = "Proceedings of the 54th Annual Meeting of the Association for Computational Linguistics",
  year = "2016",
  pages = "225--230"
}

@inproceedings{Zhang:EMNLP:2020,
  title = "Multi-modal Multi-label Emotion Detection with Modality and Label Dependence",
  author = "Zhang, Dong and Ju, Xincheng and Li, Junhui and Li, Shoushan and Zhu, Qiaoming and Zhou, Guodong",
  booktitle = "Proceedings of EMNLP 2020",
  year = "2020",
  pages = "3584--3593"
}

@article{Charte:Neuro:2015,
  title={Addressing imbalance in multilabel classification: Measures and random resampling algorithms},
  author={Charte, Francisco and Rivera, Antonio J. and Del Jesus, María J. and Herrera, Francisco},
  journal={Neurocomputing},
  volume={163},
  pages={3--16},
  year={2015}
}

@article{Bach:IS:2017,
  title={The study of under-and over-sampling methods’ utility in analysis of highly imbalanced data on osteoporosis},
  author={Bach, Małgorzata and Werner, Aleksandra and Żywiec, J. and Pluskiewicz, Wojciech},
  journal={Information Sciences},
  volume={384},
  pages={174--190},
  year={2017}
}

@misc{Muhammad:BRIGHTER:2025,
  title={BRIGHTER: BRIdging the Gap in Human-Annotated Textual Emotion Recognition Datasets for 28 Languages},
  author={Muhammad, Shamsuddeen Hassan and Ousidhoum, Nedjma and Abdulmumin, Idris and Wahle, Jan Philip and Ruas, Terry and Beloucif, Meriem and De Kock, Christine and Surange, Nirmal and Teodorescu, Daniela and Ahmad, Ibrahim Said and Adelani, David Ifeoluwa and Aji, Alham Fikri and Ali, Felermino D. M. A. and Alimova, Ilseyar and Araujo, Vladimir and Babakov, Nikolay and Baes, Naomi and Bucur, Ana-Maria and Bukula, Andiswa and Cao, Guanqun and Cardenas, Rodrigo Tufino and Chevi, Rendi and Chukwuneke, Chiamaka Ijeoma and Ciobotaru, Alexandra and Dementieva, Daryna and Gadanya, Murja Sani and Geislinger, Robert and Gipp, Bela and Hourrane, Oumaima and Ignat, Oana and Lawan, Falalu Ibrahim and Mabuya, Rooweither and Mahendra, Rahmad and Marivate, Vukosi and Piper, Andrew and Panchenko, Alexander and Porto Ferreira, Charles Henrique and Protasov, Vitaly and Rutunda, Samuel and Shrivastava, Manish and Udrea, Aura Cristina and Wanzare, Lilian Diana Awuor and Wu, Sophie and Wunderlich, Florian Valentin and Zhafran, Hanif Muhammad and Zhang, Tianhui and Zhou, Yi and Mohammad, Saif M.},
  year={2025}
}

@Inbook{Tsoumakas:DKDH:2010,
  author={Tsoumakas, Grigorios and Katakis, Ioannis and Vlahavas, Ioannis},
  title="Mining Multi-label Data",
  bookTitle="Data Mining and Knowledge Discovery Handbook",
  year="2010",
  publisher="Springer US",
  pages="667--685"
}

@inproceedings{Yang:ACL:2019,
  title = "A Deep Reinforced Sequence-to-Set Model for Multi-Label Classification",
  author = "Yang, Pengcheng and Luo, Fuli and Ma, Shuming and Lin, Junyang and Sun, Xu",
  booktitle = "Proceedings of ACL 2019",
  year = "2019",
  pages = "5252--5258"
}

@inproceedings{Ribeiro:ACL:2020,
  title={Beyond Accuracy: Behavioral Testing of NLP Models with CheckList},
  author={Ribeiro, Marco Tulio and Singh, Sameer and Guestrin, Carlos},
  booktitle={Proceedings of ACL 2020},
  pages={4904--4913},
  year={2020}
}

@article{Loshchilov:AdamW:2017,
  title={Decoupled weight decay regularization},
  author={Loshchilov, Ilya and Hutter, Frank},
  journal={arXiv preprint arXiv:1711.05101},
  year={2017}
}

@INPROCEEDINGS{Ridnik:ICCV:2021,
  author={Ridnik, Tal and Ben-Baruch, Emanuel and Zamir, Nadav and Noy, Asaf and Friedman, Itamar and Protter, Matan and Zelnik-Manor, Lihi},
  title={Asymmetric Loss For Multi-Label Classification},
  booktitle={Proceedings of ICCV 2021},
  year={2021},
  pages={82--91}
}
